\documentclass{article}
\usepackage{arxiv}
\usepackage{makecell}
\usepackage{longtable}
\usepackage{graphicx}
\usepackage{amsmath}
\usepackage{url}
\usepackage{hyperref}

\usepackage[T1]{fontenc}
\usepackage[utf8]{inputenc}
\usepackage[english]{babel}

\usepackage[authoryear]{natbib}
\title{Balancing Sustainability and Performance: The Role of Small-Scale LLMs in Agentic Artificial Intelligence Systems}
\renewcommand{\shorttitle}{Balancing Sustainability and Performance}

\date{}

\author{Anh-Khoa Ngo-Ho \\
Capgemini Invent, France \\
\texttt{anh-khoa.ngo-ho@capgemini.com} \\
\And
Martin Chauvin \\
Capgemini Invent, France \\
\And
Simon Gosset \\
Capgemini Invent, France \\
\And
Philippe Cordier \\
Capgemini Invent, France \\
\And
Boris Gamazaychikov \\
Salesforce, France \\
}

\begin{document}
\maketitle
\begin{abstract}
As large language models become integral to agentic artificial intelligence systems, their energy demands during inference may pose significant sustainability challenges. This study investigates whether deploying smaller-scale language models can reduce energy consumption without compromising responsiveness and output quality in a multi-agent, real-world environments. We conduct a comparative analysis across language models of varying scales to quantify trade-offs between efficiency and performance. Results show that smaller open-weights models can lower energy usage while preserving task quality. Building on these findings, we propose practical guidelines for sustainable artificial intelligence design, including optimal batch size configuration and computation resource allocation. These insights offer actionable strategies for developing scalable, environmentally responsible artificial intelligence systems.
\end{abstract}

%% JAIR Note: 
%% Do not include ACM CCS Concepts or Keywords

%% To be updated by authors.
%\received{20 February 2007}
%\received[accepted]{5 June 2009}

%%
%% This command processes the author and affiliation and title
%% information and builds the first part of the formatted document.
\maketitle

\section{Introduction}
The integration of large language models (LLMs) into agentic artificial intelligence (AI) systems is rapidly expanding, driven by the demand for intelligent automation, personalized interactions, and AI-driven decision-making. As LLMs become central to multi-agent architectures, concerns about their environmental footprint, particularly during inference, are gaining prominence \citep{maliakel_investigating_2025}. Recent work by \citet{desroches_exploring_2025} reports a quadratic relationship between the adoption rate of agentic AI systems and associated energy consumption. Building on this context, our study explores the sustainability implications of LLM size in the deployment of AI agents at scale. Specifically, it investigates how model size reduction can mitigate energy consumption while keeping acceptable levels of performance and user experience.

This research navigates a tri-objective optimization space defined by \textbf{Environmental Impact}, \textbf{User Experience}, and \textbf{Output Quality}. First, \textbf{Environmental Impact} quantifies the computational energy savings realized by transitioning to smaller-scale LLMs. Second, \textbf{User Experience} evaluates latency, determining if reduced model architectures can maintain, or improve, responsiveness in agentic systems. Finally, \textbf{Output Quality} measures the degree to which smaller models preserve the semantic accuracy and reliability essential for large-scale enterprise applications.

To the best of our knowledge, few studies have comprehensively examined the trade-offs between all three objectives across different model sizes and compression strategies, particularly in real-world deployment scenarios. Most existing research addresses only partial aspects of this multi-objective trade-off. Some studies focus exclusively on a single dimension, such as output quality \citep{lee_exploring_2025}, user experience metrics like latency or token generation speed \citep{agrawal_evaluating_2025}, or energy consumption \citep{luccioni_power_2024, chung_mlenergy_2025, jegham_how_2025}. Others explore pairwise trade-offs, for example, accuracy versus latency or throughput \citep{kurtic_give_2025}, or energy efficiency versus accuracy \citep{reus_exploration_2024, jin_energy_2025}. However, only a limited number of works attempt to jointly analyze all three dimensions \citep{yang_llmcbench_2024, shi_systematic_2025, maliakel_investigating_2025}. Notably, these works primarily evaluate LLM performance in controlled research settings using curated datasets, rather than in real-world deployment environments.
Additionally, the scope of these studies is often constrained. Some focus solely on quantized models \citep{yang_llmcbench_2024, shi_systematic_2025}, while others limit their analysis to specific model sizes or families \citep{nguyen_towards_2024, jin_energy_2025, maliakel_investigating_2025}. In contrast, our work aims to provide a more holistic perspective by evaluating the trade-offs among output quality, user experience, and environmental impact across a diverse set of LLMs (varying in size, architecture, and the application of compression techniques).
To address these objectives and bridge the identified research gap, we conducted a comparative analysis of open-weights LLMs of varying sizes against a baseline closed-source model commonly used in real-world deployments (i.e., GPT-4o). Evaluations were performed under conditions representative of real-world AI agent deployments. This experimental design enables a critical assessment of whether smaller-scale open-weights models can realistically challenge the dominance of closed-source LLMs. Furthermore, the proposed method is replicable for conducting similar analyses across other AI systems. Our findings offer actionable insights into the trade-offs between model efficiency and performance, contributing to the development of more sustainable, scalable, and responsible AI systems for enterprise use. 

\section{Related Works}
In the case of energy consumption, \citet{lacoste_quantifying_2019} evaluated the CO2-equivalent emissions during the training phase of a machine learning model, by considering several factors: the server's location and the energy grid it utilizes, the duration of the training process, and model of the hardware used. Building on this work, \citet{lannelongue_green_2020} developed a freely available online tool that also considers the memory requested. Additionally, \citet{henderson_towards_2022} provided a lightweight framework for easily tracking the energy usage and carbon impact of a machine learning model. Another well-known offline tracker is CodeCarbon \citep{courty_mlco2codecarbon_2024}, which measures the energy consumption and carbon footprint of CPUs, GPUs, and RAM. Cloud Carbon Footprint \citep{thoughtworks_inc_cloud_2020} is an attempt that calculates CO2 emissions using usage data (e.g., compute, storage, networking, etc.) from popular cloud providers such as AWS, Google Cloud, and Microsoft Azure. Several studies have directly measured energy consumption and prompt-level metrics on AI accelerator hardware, including \citet{luccioni_estimating_2022, samsi_words_2023, luccioni_ai_2025, chung_mlenergy_2025}. Several studies have attempted to estimate the energy consumption of closed-source models, including those by \citet{jegham_how_2025, desroches_exploring_2025}. Moreover, there are several initiatives to collect the energy consumption of AI models, such as LLM-Perf Leaderboard\footnote{\url{https://huggingface.co/spaces/optimum/llm-perf-leaderboard}}, ML.ENERGY Leaderboard\footnote{\url{https://ml.energy/leaderboard}} and AI Energy Score\footnote{\url{https://huggingface.co/spaces/AIEnergyScore/Leaderboard}}. \citet{chung_mlenergy_2025} introduced ML-Energy Benchmark, a benchmarking framework designed to evaluate the intricate balance between energy usage, latency, and model architecture. Their method replicates realistic API-based deployment conditions by monitoring energy consumption for each individual or batched request, both on the user's device and the remote server hosting the large-scale models. Additionally, the study highlights that batch size is a crucial configuration parameter, as it has a significant impact on both generation latency and energy consumption. Given its ability to simulate real-world deployment conditions for LLMs, we adopt this framework as the foundation of our study.
\citet{maliakel_investigating_2025} explored the energy-performance trade-offs of LLM inference, highlighting the impact of input complexity and hardware-level optimizations. By analyzing prompt features and GPU configurations, the study identifies strategies (e.g., Dynamic Voltage and Frequency Scaling, supported by modern GPUs like the NVIDIA A100) that can reduce energy consumption by up to 30\% without degrading output quality, contributing to more sustainable LLM deployment.
To mitigate the large computational and parameter overhead associated with LLMs, a variety of model compression techniques have been developed in recent years. These methods focus on three areas: sparsification, quantization, and knowledge distillation \citep{yang_llmcbench_2024}. Model sparsification is a technique designed to drop redundant weights or activations to produce a sparse model, thereby reducing both the parameter count and computational load of LLMs. Quantization involves reducing the bit precision of model parameters to lower computational requirements while keeping acceptable inference accuracy. Knowledge Distillation is a method used to transfer learned representations from a large, complex model (the teacher) to a smaller, more efficient model (the student). 
\citet{yang_llmcbench_2024} provided a fair benchmark framework LLMCBench for compression methods by tracking accuracy (via compression performance, generalization ability, model trustworthiness), user experience and environmental impact (via training-inference consumption and hardware acceleration). It evaluates the two well-known compression methods (model sparsification and quantization) applied for four model families including LLaMA, Vicuna, OPT, and ChatGLM include different model sizes ranging from 6B to 70B. Recently, \citet{shi_systematic_2025} analyzed the trade-offs between latency, energy, and accuracy in quantization methods for Llama models. They concluded that no single quantization method excels across all metrics and performance varies by task, model size, and precision.

\section{Benchmark}
\subsection{Evaluated Agent}
\begin{figure}[ht]
  \centering
  \includegraphics[width=0.8\linewidth]{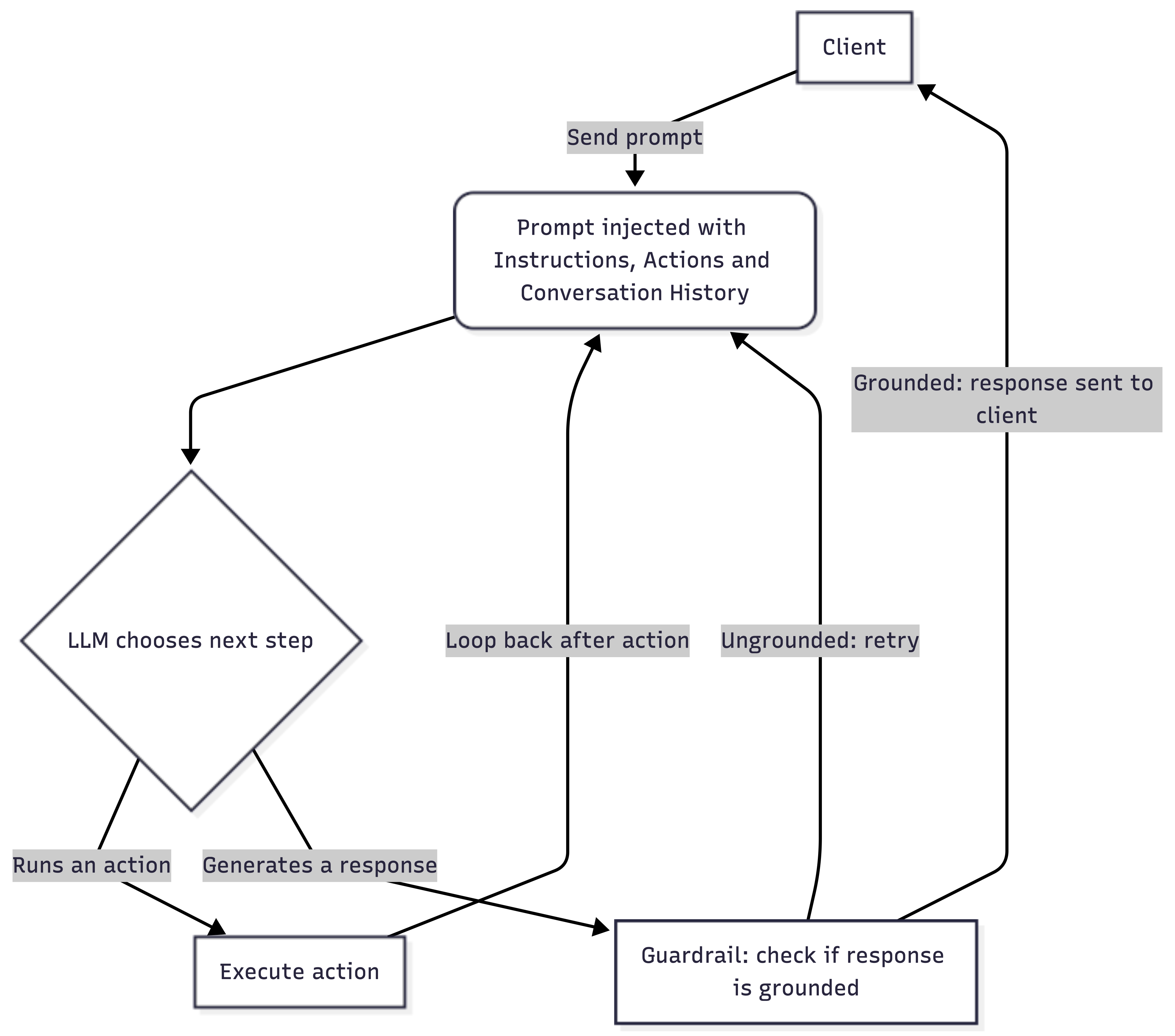}
  \caption{Workflow of a multi-agent framework interacting with a client.}
%  \Description{Workflow of a multi-agent framework interacting with a client.}
  \label{fig:workflow}
\end{figure}

Our analysis centers on the reasoning engine embedded within a real-world multi-agent framework, which orchestrates the behavior of LLM-based agents in response to end user interactions. Figure \ref{fig:workflow} illustrates a simplified workflow of the framework interacting with a client. It starts when the client sends a prompt, which is enriched with instructions, actions, and conversation history. The LLM then decides the next step: either generating a response or executing an action. After generating a response, a validation mechanism checks if the response is grounded (accurate and appropriate). If grounded, the response is sent back to the client; if ungrounded, the system retries. This mechanism is designed to uphold system integrity and mitigate hallucinations which is an inherent challenge in LLM systems.

In this study, we evaluate this validation mechanism which decides whether a generated response is contextually grounded or hallucinated. This component works using a tailored prompt to classify responses as grounded, ungrounded, or small talk. It receives a structured JSON input including the context, conversation history, and the generated response, and returns a JSON output containing a classification label (“grounded”, “ungrounded”, or “small talk”), justification, and supporting evidence drawn from the input context and history (see Table \ref{tab:json_output}). We additionally incorporate outputs from a real-world deployment of the agent, powered by the closed-source GPT-4o model. This allows us to assess behavioral changes in the agent when substituting the closed-source model with smaller-scale open-weights alternatives.

\begin{table}
	\centering
	\caption{Examples of JSON outputs evaluated by the reference models}
	\label{tab:json_output}
	\begin{tabular}{|p{0.8\linewidth}|}
    	\hline
\{
"result": "GROUNDED",\\
"sources": ["context[2]", "functions[1]"],\\
"reason": "A function was called to verify a refund is 
allowed. The claim in the response about the order being eligible for return is grounded."
\} \\ \\
\{
"result": "UNGROUNDED",
"sources": [],
"reason": "The response was not grounded in the resources because the claim about the 
policy is not supported by the context, 
function\_history or conversation\_history."
\} \\
	\hline
	\end{tabular}
\end{table}

\subsection{Benchmarking Metrics}
Based on the objectives mentioned above, we define a set of metrics to evaluate our benchmark across three key dimensions: environmental impact, user experience, and output quality.

\paragraph{\textbf{Environmental Impact} via \textbf{Energy Consumption Per Request}:} We quantify the average energy consumption per request on GPU s, measured in joules, using the ML-Energy Benchmark \citep{chung_mlenergy_2025}. This measurement reflects energy usage under steady-state and production-like conditions, where GPU use is maximized.
\paragraph{\textbf{User Experience} via \textbf{Decode Latency Per Request}:} User experience is evaluated through latency, which reflects both system responsiveness and perceived service speed from the user's perspective \citep{huang_latency_2025}. Inference includes both the prefill phase (i.e., prompt processing) and the decode phase (i.e., token generation) \citep{patel_splitwise_2024}. In this study, we focus on the duration of the decode phase because, in the multi-agent system, the validation mechanism must complete its response before subsequent tasks can proceed, making this metric a strong indicator of overall service speed. Prefill latency primarily captures initial responsiveness, which is more critical when users expect immediate feedback (i.e., the first generated token) \citep{wang_prefill-decode_2025}. Additionally, decode latency correlates with output length, and our initial experiments revealed substantial variation in the number of generated tokens across models.
\paragraph{\textbf{Output quality}:} We evaluate output quality using two complementary approaches.
\begin{itemize}
\item \textbf{F1-score}: The agent's output includes a classification label (“grounded”, “ungrounded”, or “small talk”), along with justification and supporting evidence from the conversation context, formatted in JSON. The classification label is the most critical component.
To assess whether the evaluated LLMs produce correctly formatted outputs with accurate labels, we compute the F1-score. If outputs are JSON-unserializable or contain labels outside the predefined set, they are marked as “Error”.
Reference labels are derived from a real-world deployment using the GPT-4o model, enabling direct comparison with smaller-scale open-weights alternatives. In this context, the F1-score quantifies the alignment between the classification labels predicted by the open-weights LLMs and those produced by the reference model, thereby capturing discrepancies in prediction accuracy.
The reference model assigns 916 samples as “grounded”, 20 samples as “ungrounded”, and 64 samples as “small talk”. Since the dataset exhibits class imbalance and a bias toward the majority class, we use the Macro F1-score for evaluation.

\item \textbf{LLM-as-a-Judge}: We also use a LLM as an independent evaluator to assess the general quality of responses from s maller-scale models. For this purpose, we adopt the Rubric-Based Criteria Scoring Metric implemented in Ragas \citep{es_ragas_2025}, where an LLM (i.e., GPT-5) assigns scores based on predefined descriptions. We define three scoring levels:
	\begin{enumerate}
	\item The response is factually incorrect, misleading, or irrelevant to the input.
	\item The response is partially correct or relevant but contains notable inaccuracies, ambiguities, or lacks clarity.
	\item The response is factually accurate, relevant, and clearly addresses the input without significant issues.
	\end{enumerate}
	Finally, all scores are normalized to a 0-1 scale to ensure consistency across evaluations, where 0 corresponds to level (1), 0.5 to level (2), and 1 to level (3).
\end{itemize}

We follow the approach of \citep{yang_llmcbench_2024} to compute the Overall Metric (OM), which provides a unified score for ranking alternatives to the reference model (GPT-4o) across three dimensions: output quality, decode latency, and energy consumption. The metric is defined as:

\begin{equation}
  \operatorname{OM} = \sqrt{w_{\operatorname{Quality}}(\frac{\operatorname{Quality}^E}{\operatorname{Quality}^R})^2 + w_{\operatorname{Latency}}(\frac{\operatorname{Latency}^R}{\operatorname{Latency}^E})^2 + w_{\operatorname{Energy}}(\frac{\operatorname{Energy}^R}{\operatorname{Energy}^E})^2 }
\label{eq:overall_metric}
\end{equation}

where $w_{\operatorname{Quality}}$, $w_{\operatorname{Latency}}$, $w_{\operatorname{Energy}}$ are the importance weights assigned to each dimension, summing to 1, and reflecting their relative significance in the evaluation. Here, E denotes the evaluated model, while R refers to the reference model. For the quality objective, we use LLM-as-a-Judge as the evaluation metric. A higher OM shows better overall performance, and values greater than one signify performance superior to the reference model.

\section{Experiment Settings}
\subsection{Evaluated LLMs}
In our experiments, we evaluated 28 open-weights LLMs\footnote{All open-weights LLMs used in our experiments were sourced from the Hugging Face Model Hub.} with 7 compressed models to understand their behavior across different model sizes. Specifically, we examined the Qwen 2.5 Instruct series (one of the most size-diverse model families available) which includes seven models with parameter sizes of 0.5B, 1.5B, 3B, 7B, 14B, 32B, and 72B \citep{yang_qwen2_2024}. This allowed us to analyze performance trends within a single model family as model size varies.
To investigate the impact of model compression, we focused on the Qwen 2.5 7B model and applied two common compression techniques: quantization and knowledge distillation (KD). For quantization, we utilized the publicly released versions of Qwen2.5-7B-Instruct with Activation-aware Weight Quantization (AWQ) \citep{lin_awq_2024}, as well as post-training quantized variants using GPTQ in 4-bit and 8-bit formats \citep{frantar_sparsegpt_2023}. For knowledge distillation, we evaluated the DeepSeek-R1-Distill-Qwen models \citep{deepseek-ai_deepseek-r1_2025} at 1.5B and 7B parameter scales. These experiments allow us to assess whether compression techniques can effectively reduce energy consumption while maintaining acceptable output quality. Because the performance of these techniques is highly sensitive to hyperparameter choices, we primarily relied on publicly released configurations. This approach reflects real-world deployment practices, where pre-trained and well-tested versions are often preferred over custom tuning.
To explore the feasibility of replacing the reference model (i.e., the closed-source GPT-4o model) used in real-world deployments, we extended the evaluation to include 20 additional LLMs. These include:

\begin{itemize}
\item Qwen 3: Dense models (4B-Instruct-2507, 8B, 14B, 32B) and a mixture-of-experts (MoE) model (30B-A3B)
\item Distilled variants of Qwen using DeepSeek-R1 \citep{deepseek-ai_deepseek-r1_2025}: DeepSeek-R1-Distill-Qwen-14B and DeepSeek-R1-0528-Qwen3-8B.
\item Gemma 3: Instruction-tuned models with 4B, 12B, and 27B parameters.
\item Mistral: Instruction-tuned models including Nemo-Instruct-2407 (12B parameters), Small-24B-Instruct-2501 (24B parameters), and Large-Instruct-2411 (123B parameters).
\item Falcon 3: Instruction-tuned LLMs with 1B, 3B, 7B and 10B parameters.
\item Phi 4: Version of reasoning and reasoning-plus (14B).
\item Llama-4-Scout-17B-16E-Instruct: A MoE model including 16 experts with 17B activated parameters, resulting in a total of approximately 109B parameters.
\end{itemize}

\subsection{Benchmark Dataset}
We evaluated these LLMs using prompts extracted from a representative sample of conversations within a multi-agent system deployed in a real-world application. The evaluation dataset consists of 1,000 requests, each averaging approximately 8,000 tokens, with the longest request reaching up to 25,500 tokens. Using the reference model GPT-4o, accessed via API, the evaluated agent produced responses averaging 66 tokens in length, with a maximum response length of 118 tokens. All experimental results are publicly available at \url{https://doi.org/10.5281/zenodo.17802868}.

\subsection{Evaluation Environment}
We conducted our experiments using the ML-Energy Benchmark framework\footnote{We reused the code corresponding to the October 16, 2024 version available at \url{https://github.com/ml-energy/leaderboard/commits/master/benchmark/llm\_text\_generation/chat}.} \citep{chung_mlenergy_2025}, which enables measurement of inference time and energy consumption across different model configurations. The framework integrates Zeus\footnote{\url{https://ml.energy/zeus/}}, a programmatic energy measurement library, to collect energy usage data. Since individual requests often overlap due to iteration-level batching, direct per-request energy attribution is not feasible. To address this, the framework identifies a steady state (i.e., a period during which the server operates at its maximum batch size) to approximate real-world deployment conditions. During this phase, the average energy consumption per token is computed and used to estimate per-request energy usage based on the number of output tokens. Building on the findings of \citet{chung_mlenergy_2025} which emphasize batch size as a key factor influencing both decode latency and energy consumption, we evaluated our LLMs under varying maximum batch sizes. Specifically, we experimented with maximum batch sizes of 2, 4, 8, 16, 32, 128, 256, and 512 to assess their impact on performance and efficiency.

\begin{table}
	\centering
	\caption{Number of GPUs Needed to Deploy Each Large Language Model}
	\label{tab:num_gpu}
	\begin{tabular}{|c|p{0.6\linewidth}|}
    		\hline
		Number of GPUs & LLMs \\
		\hline \hline
		1 & Nano LLMs. Qwen2.5-Instruct (0.5B, 1.5B, 3B, 7B), Qwen3 (4B-Instruct-2507, 8B), DeepSeek-R1-Distill-Qwen (1.5B, 7B), DeepSeek-R1-0528-Qwen3-8B, gemma-3-4b-it, Falcon3-Instruct (1B, 3B, 7B, 10B) \\
		\hline
		2 & Micro LLMs. Qwen2.5-14B-Instruct, Qwen3-14B, DeepSeek-R1-Distill-Qwen-14B, gemma-3-12b-it, Mistral-Nemo-Instruct-2407, Phi-4-reasoning and its -plus version\\
		\hline
		3 & Small LLMs. Qwen2.5-32B-Instruct, Qwen3-32B, Qwen3-30B-A3B-Instruct-2507, gemma-3-27b-it, Mistral-Small-24B-Instruct-2501 \\
		\hline
		4 & Large LLMs. Qwen2.5-72B-Instruct, Mistral-Large-Instruct-2411, Llama-4-Scout-17B-16E-Instruct \\
		\hline
	\end{tabular}
\end{table}

The framework uses vLLM v0.5.4\footnote{\url{https://github.com/vllm-project/vllm}}, a high-performance inference and serving engine that supports most popular open-weights models hosted on the Hugging Face platform\footnote{\url{https://huggingface.co}}. To accommodate recently released models such as Qwen 3 and Gemma 3, we upgraded the vLLM version and modified the benchmark codebase\footnote{Following the ML-Energy Benchmark framework, we integrated Zeus into the upgraded vLLM codebase to measure steady-state energy consumption.}. For the reference model GPT-4o, accessed via API, we estimated energy consumption and decode latency based on the methodology proposed by \citet{jegham_how_2025}, due to the lack of direct access to hardware-level measurements. 
All experiments were conducted on an AWS p4d.24xlarge instance\footnote{\url{https://aws.amazon.com/fr/ec2/instance-types/p4/}}, equipped with 8 NVIDIA A100 Tensor Core GPUs, each with 40 GB of memory. This instance type is widely recognized as a standard for large-scale LLM experimentation. Drawing inspiration from hardware classifications based on model size \citep{jegham_how_2025}, we grouped the evaluated LLMs into four distinct hardware classes. Table \ref{tab:num_gpu} summarizes the number of GPUs used for each hardware class.

\subsection{Result and Discussion}
\subsection{Open-Weights LLM Alternatives for Reducing Environmental Impact}

\begin{table}
	\centering
	\caption{Energy Consumption, Decode Latency, and Output Quality (per request) of Open-Weights LLMs
Compared to GPT-4o. Scores shown for the lowest-energy configuration among tested batch sizes. The best values for each metric among the open-weights LLMs are highlighted in \textbf{bold}.}
	\label{tab:result}
	\begin{tabular}{|p{0.3\linewidth}||p{0.15\linewidth}|p{0.15\linewidth}|p{0.15\linewidth}|p{0.1\linewidth}|}
    	\hline
		\textbf{Models} & \textbf{Energy} (Joules) & \textbf{Decode Latency} (Secs) & \multicolumn{2}{c|}{\textbf{Output Quality}} \\
		& & & \textbf{LLM-as-a-Judge} & \textbf{F1-score} \\
		\hline \hline
		Baseline: gpt-4o-2024-05-13 & 1499 $\pm$ 287 & \textbf{0.58 $\pm$ 0.1} & 0.978 $\pm$ 0.116 & 1.0 \\
		Qwen3-32B & 2382 $\pm$ 725 & 85 $\pm$ 26 & \textbf{0.992 $\pm$ 0.064} & \textbf{0.918} \\
		Qwen3-14B & 1177 $\pm$ 820 & 71 $\pm$ 49 & 0.981 $\pm$ 0.101 & 0.876 \\
		Qwen3-30B-A3B-Instruct-2507 & 456 $\pm$ 97 & 50 $\pm$ 10 & 0.956 $\pm$ 0.156 & 0.917 \\
		DeepSeek-R1-0528-Qwen3-8B & 769 $\pm$ 640 & 56 $\pm$ 47 & 0.962 $\pm$ 0.143 & 0.862 \\
		DeepSeek-R1-Distill-Qwen-14B & 1104 $\pm$ 296 & 55 $\pm$ 14 & 0.955 $\pm$ 0.160 & 0.854 \\
		Mistral-Large-Instruct-2411 & 8281 $\pm$ 2199 & 21 $\pm$ 5 & 0.958 $\pm$ 0.165 & 0.786 \\
		Mistral-Nemo-Instruct-2407 & 534 $\pm$ 358 & 24 $\pm$ 16 & 0.840 $\pm$ 0.301 & 0.747 \\
		Falcon3-10B-Instruct & 458 $\pm$ 305 & 11 $\pm$ 7 & 0.905 $\pm$ 0.246 & 0.773 \\
		Falcon3-7B-Instruct & \textbf{335 $\pm$ 348} & 17 $\pm$ 17 & 0.885 $\pm$ 0.245 & 0.910 \\
		Phi-4-reasoning & 2094 $\pm$ 1618 & 117 $\pm$ 90 & 0.975 $\pm$ 0.122 & 0.488 \\
		\hline
	\end{tabular}
\end{table}

\begin{figure}[ht]
  \centering
  \includegraphics[width=0.9\linewidth]{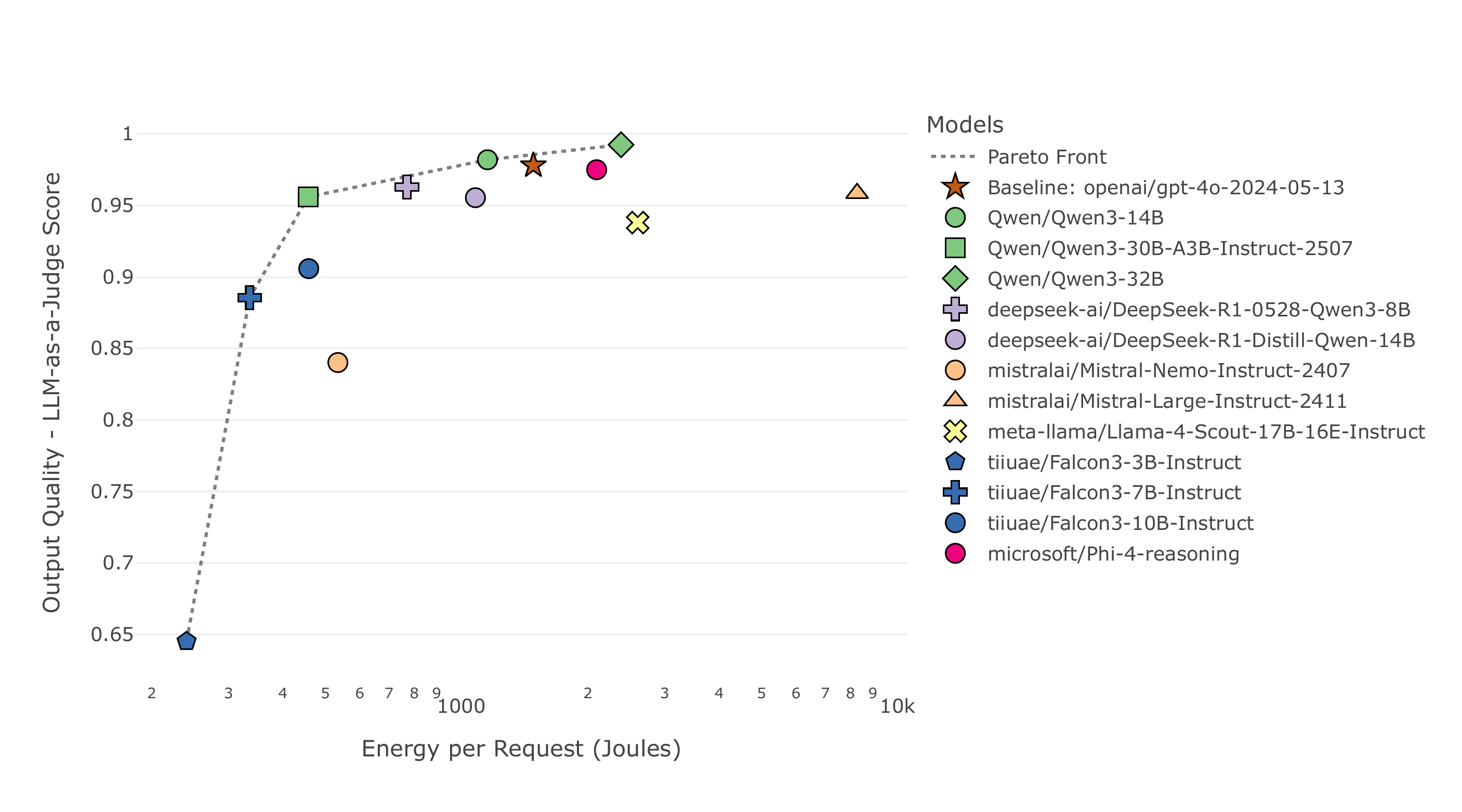}
  \caption{Output Quality versus Energy Consumption of the Best-performing Open-Weights LLMs.
Scores shown for the lowest-energy configuration among tested batch sizes. Colors denote LLM families, while symbols represent model sizes. To simplify visualization, a single symbol may correspond to multiple closely related sizes (e.g., circles for 10B, 12B, and 14B; crosses for 7B and 8B).}
%\Description{Output Quality versus Energy Consumption of the Best-performing Open-Weights LLMs. Scores shown for the lowest-energy configuration among tested batch sizes. Colors denote LLM families, while symbols represent model sizes. To simplify visualization, a single symbol may correspond to multiple closely related sizes (e.g., circles for 10B, 12B, and 14B; crosses for 7B and 8B).}
  \label{fig:energy_quality}
\end{figure}

Table \ref{tab:result} displays the output quality (measured by F1-score and LLM-as-a-Judge\footnote{In certain cases, the sum of the mean and the standard deviation of LLM-as-a-Judge exceeds 1 because the distribution is significantly skewed to the right.}) alongside decode latency and energy consumption per request for the ten best-performing open-weights LLMs against the reference model, GPT-4o. Model selection was driven by both output quality scores, and the reported scores correspond to the configuration with the lowest energy consumption among the tested maximum batch sizes (see Table \ref{tab:result_gpu} in Annex). As previously noted, the energy consumption and decode latency values for GPT-4o are manually estimated based on the methodology from \citet{jegham_how_2025}\footnote{Decode Latency for GPT-4o is estimated as number of output tokens $\div 113.7 tokens/s$; energy consumption (Joules) is calculated as $7.7 kW \times (0.5 + decode latency) \times 0.18 PUE \times 1000$, where $7.7 kW$ is GPU power, $0.18$ is Power Usage Effectiveness, and $0.5$ seconds is the average prefill duration. Values are based on Artificial Analysis (\url{https://artificialanalysis.ai}).}. Among the evaluated models, the Qwen3 family consistently delivers the highest performance. Qwen3-30B-A3B-Instruct-2507 (a mixture-of-experts model) and Qwen3-32B achieve output quality most comparable to GPT-4o, with only a 10\% reduction in F1-score. Interestingly, the LLM-as-a-Judge metric assigns even higher scores to Qwen3-32B (0.992) and Qwen3-14B (0.981) compared to GPT-4o (0.978). Substituting GPT-4o with Qwen-3-30B-A3B-Instruct-2507 (-2.2\% in LLM-as-a-Judge) offers a substantial efficiency gain, achieving a 70\% reduction in energy consumption per request. This improvement stems from its mixture-of-experts architecture, which activates just 3.3B parameters during inference. Despite the smaller active parameter count, the quality reduction remains minimal. This can be attributed to the model’s fine-grained expert segmentation and the global-batch load balancing loss applied during training \citep{yang_qwen3_2025}. Figure \ref{fig:energy_quality} further illustrates the Pareto front, highlighting Qwen3 models and Falcon (3B, 7B) as optimal for balancing performance and efficiency. Beyond a certain point (near Qwen3-30B-A3B-Instruct-2507) additional energy expenditure yields minimal gains in output quality. \textbf{Therefore, substituting a closed-source model with an open-weights alternative is feasible following careful evaluation, as it can substantially reduce energy consumption with minimal impact on output quality.}

The decline in output quality is primarily due to two factors: (1) the models occasionally fail to generate correctly formatted JSON outputs, and (2) they sometimes include additional explanations not required by the task. For decode latency, the reference models accessed via API show the shortest response times. One contributing factor is that the evaluated models often generate longer outputs compared to the reference model. This behavior may be influenced by the fact that we reused prompts optimized for GPT-4o, the reference model, which could lead to inefficiencies when applied to smaller-scale alternatives. Mitigating these issues may require prompt tuning and/or fine-tuning each model.
Additionally, decode latency may be influenced by the inference framework. In this study, we used vLLM to host all evaluated LLMs locally with default parameters. A potential improvement is to fine-tune vLLM hyperparameters. Exploring alternative frameworks (e.g., TensorRT-LLM\footnote{\url{https://github.com/NVIDIA/TensorRT-LLM}}, llama.cpp\footnote{\url{https://github.com/ggml-org/llama.cpp}}) is also promising, as each employs distinct optimization strategies for inference performance. Notably, \citet{chitty-venkata_llm-inference-bench_2024} report that TensorRT-LLM delivers the highest performance and lowest power consumption on NVIDIA platforms.

\begin{figure}[ht]
  \centering
  \includegraphics[width=0.9\linewidth]{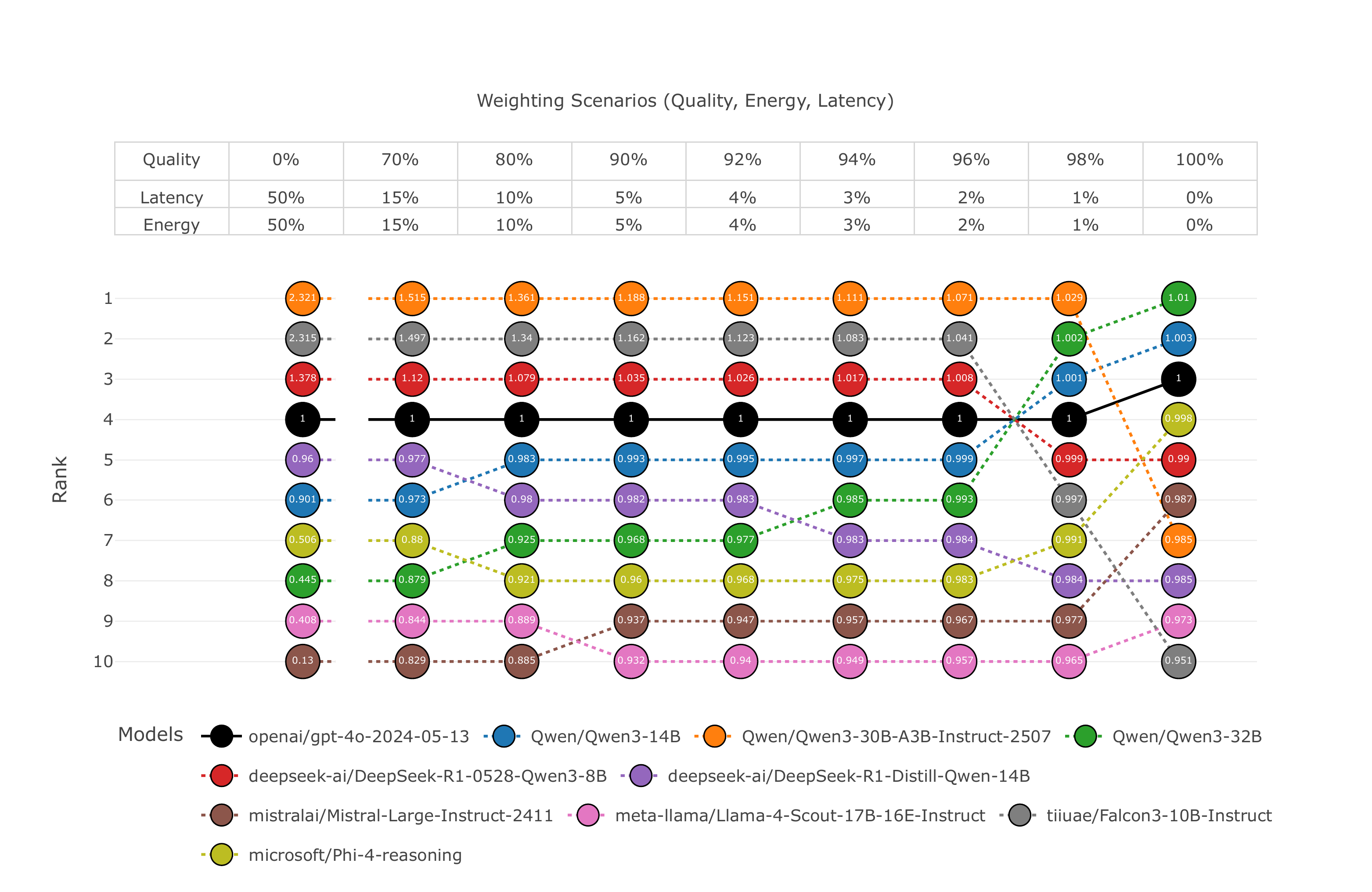}
  \caption{Ranking of models under different weighting scenarios.
The y-axis represents the rank of each model, while the table indicates the corresponding weighting scenario ($w_{Quality}$, $w_{Energy}$ and $w_{Latency}$). Node colors represent different LLMs, and the numbers inside each node indicate their overall metrics. The baseline GPT-4o is shown with solid lines, while evaluated models are shown with dotted lines.
}
%  \Description{Ranking of models under different weighting scenarios. The y-axis represents the rank of each model, while the table indicates the corresponding weighting scenario ($w_{Quality}$, $w_{Energy}$ and $w_{Latency}$). Node colors represent different LLMs, and the numbers inside each node indicate their overall metrics. The baseline GPT-4o is shown with solid lines, while evaluated models are shown with dotted lines.}
  \label{fig:ranking}
\end{figure}

We rank the alternatives to GPT-4o based on the OM under different weighting scenarios. A higher OM shows better aggregate performance, while values exceeding 1 denote superiority over GPT-4o. In Figure \ref{fig:ranking}, we vary the weight assigned to output quality to analyze its impact. Key observations include: (1) Increasing the weight on output quality reduces the score dispersion across models. This occurs because differences in decode latency and energy efficiency are generally larger than differences in output quality. (2) When decode latency and energy weights occupy at least 2\% each, rankings remain stable. The top three models (i.e., Qwen3 30B A3B Instruct 2507, Falcon3 10B Instruct, and DeepSeek R1 0528 Qwen3 8B ) consistently outperform GPT4o. As the output quality weight increases, these models’ overall scores decline, while lower-ranked models improve. This suggests  the top models excel primarily in decode latency and energy efficiency rather than output quality. (3) Ranking shifts occur only when output quality dominates ($w_{Quality}$= 100\% ). In this scenario, the previous top models lose their advantage. Qwen3 32B rises from 8th to 1st place, and Qwen3 14B moves from 6th to 2nd, followed by GPT 4o. This highlights the trade-off between efficiency (energy consumption and decode latency) and output quality. Moreover, when decode latency dominates the weighting (i.e., >90\%), GPT-4o stays the top choice; otherwise, Qwen3-30B-A3B-Instruct-2507 continues to lead.

\subsection{Scaling Behavior of Qwen 2.5: Impact of Model Size Reduction}

\begin{figure}[ht]
  \centering
  \includegraphics[width=0.9\linewidth]{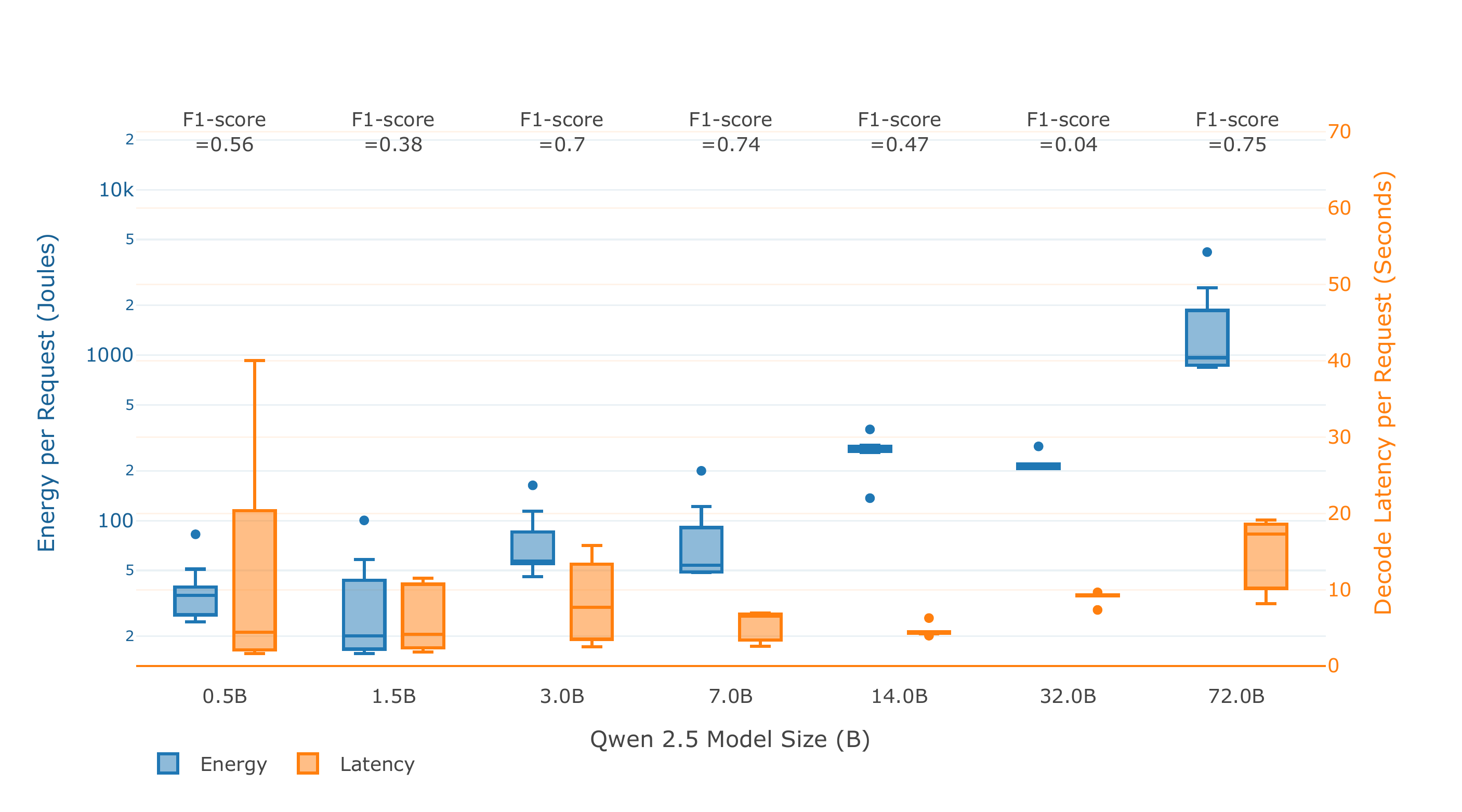}
  \caption{Energy consumption (Joules, log scale) and decode latency (seconds) per request across Qwen 2.5 model sizes (0.5B - 72B), with corresponding F1-scores shown above each group. The x-axis represents model size in billions of parameters, while the left y-axis shows energy per request (blue) and the right y-axis shows decode latency per request (orange). The variation in box sizes across models relates to differences in batch size.
}
%  \Description{Energy consumption (Joules, log scale) and decode latency (seconds) per request across Qwen 2.5 model sizes (0.5B - 72B), with corresponding F1-scores shown above each group. The x-axis represents model size in billions of parameters, while the left y-axis shows energy per request (blue) and the right y-axis shows decode latency per request (orange). The variation in box sizes across models relates to differences in batch size.}
  \label{fig:model_size}
\end{figure}

\begin{figure}[ht]
  \centering
  \includegraphics[width=0.9\linewidth]{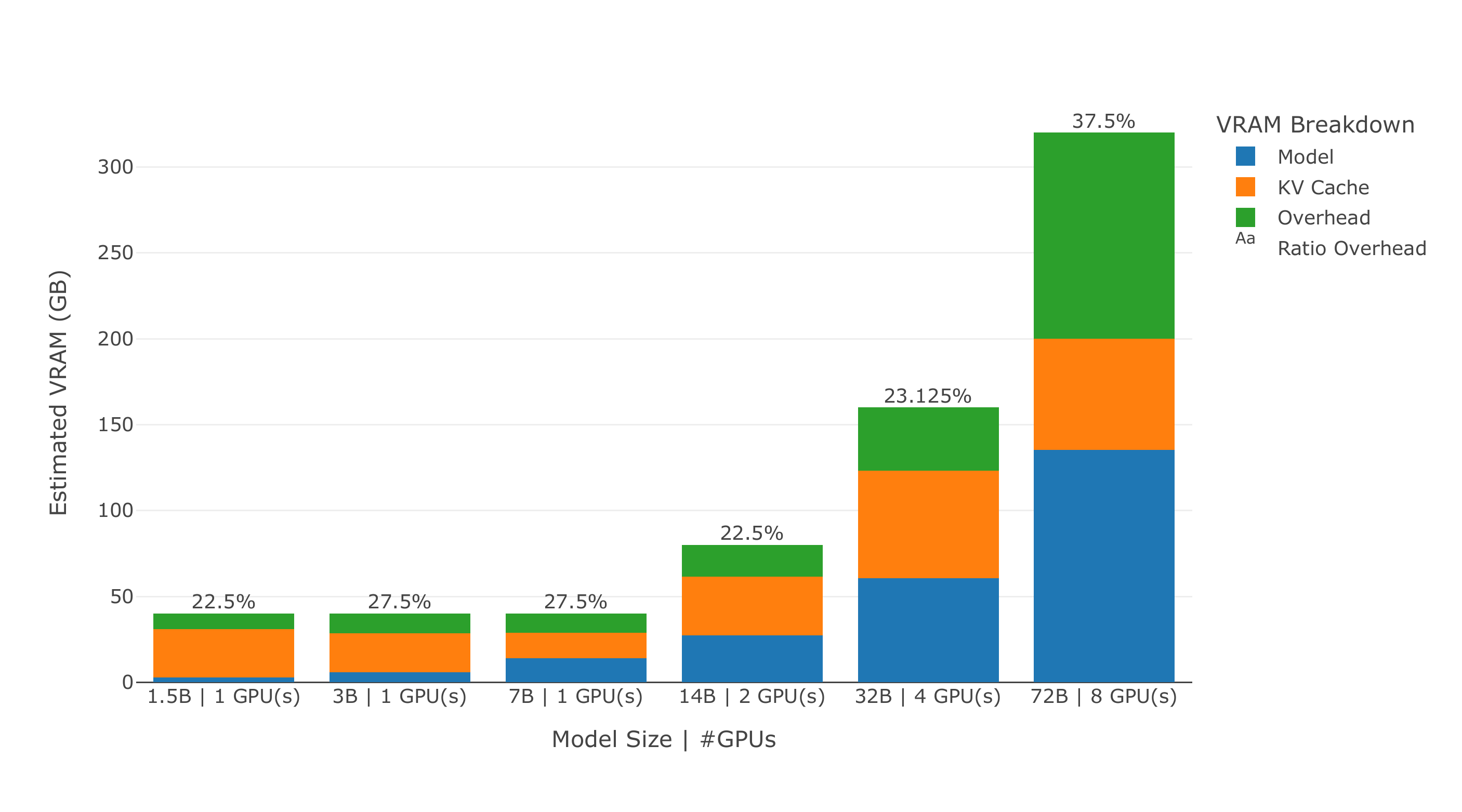}
  \caption{Estimated VRAM Requirements Across Model Sizes and GPU Configurations for Qwen 2.5. The percentage above each bar shows the ratio of overhead relative to total VRAM usage.}
%  \Description{Estimated VRAM Requirements Across Model Sizes and GPU Configurations for Qwen 2.5. The percentage above each bar shows the ratio of overhead relative to total VRAM usage.}
  \label{fig:vram}
\end{figure}

When experimenting with Qwen 2.5 models of varying sizes and varying maximum batch size, two key observations appear. First, increasing model size does not consistently lead to higher output quality. As shown in Figure \ref{fig:model_size}, the 7B parameter model achieves better output quality than the 14B and 32B models. Second, both energy consumption and decode latency increase with larger models, primarily due to the need for more GPUs. Energy consumption grows almost exponentially with model size, whereas decode latency scales nearly linearly, indicating that energy cost is more sensitive to hardware scaling than decode latency. For nano-scale models running on a single GPU (i.e., 0.5B - 3B), this upward trend is not seen. These findings suggest that using smaller models can deliver comparable output quality with substantial efficiency gains. For example, the 72B and 7B models achieve F1-scores of 0.75 and 0.74, respectively, yet the smaller model requires significantly less energy and latency because of reduced GPU memory demands.
As can be seen in Figure \ref{fig:vram}, we aim to explore energy usage through the lens of VRAM consumption. VRAM utilization can be decomposed into three components \citep{cheng_lmcache_2025}: (1) storage of model weights, (2) Key-Value (KV) cache for the input prompt, and (3) overhead associated with additional memory required for task execution and system management. Larger models and multi-GPU configurations exhibit higher memory demands, with overhead becoming increasingly significant at scale. Prior work by \citet{gond_tokenweave_2025} reports that overhead typically accounts for approximately 20\% of total VRAM usage. Our results confirm this trend and further reveal that very large models, such as the 72B parameter model, require at least 8 GPUs, resulting in overheads as high as 37\%.
For the analysis of compressed models of Qwen 2.5-7B (see Figure \ref{fig:compressed} in Annex), it is unsurprising that they achieve lower energy consumption during inference, reaffirming the findings in \citet{yang_llmcbench_2024, shi_systematic_2025}. Among quantization methods, the best-performing configuration uses GPTQ in 4-bit precision, reducing energy usage and latency by approximately 20\%, while also achieving higher output quality (F1-score) than the original full-precision model. In contrast, the AWQ method performs worst, exhibiting longer latency than the baseline. Regarding knowledge distillation, applying KD from DeepSeek-R1 to a smaller model such as 1.5B does not surpass the accuracy of the 7B model. Furthermore, KD applied to the 7B model significantly degrades performance, with an observed drop of 29\% in F1-score. Overall, compression algorithms consistently reduce energy consumption but cannot guarantee preservation of other dimensions such as output quality or decode latency.
\textbf{These observations indicate that larger-scale models do not guarantee superior performance but significantly increase energy consumption. Therefore, adopting smaller-scale models is strongly recommended as a step toward more sustainable and responsible AI.}

\subsection{Energy-Latency Trade-offs Under Varying Batch Sizes}

\begin{figure}[ht]
  \centering
  \includegraphics[width=0.9\linewidth]{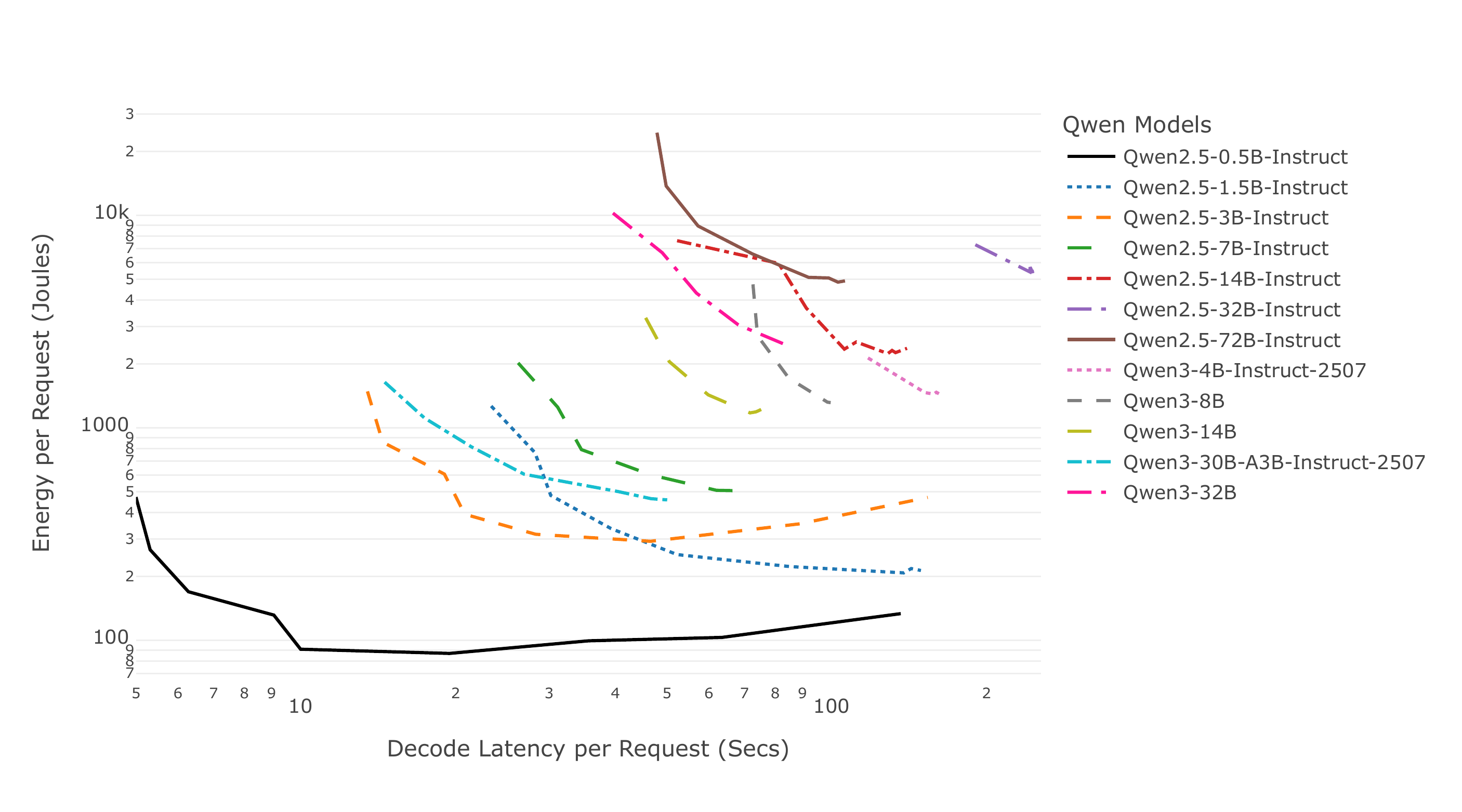}
  \caption{Energy-Latency Trade-off Across Qwen’s Model Sizes.}
%  \Description{Energy-Latency Trade-off Across Qwen’s Model Sizes.}
  \label{fig:energy_latency}
\end{figure}

By tracking energy consumption and decode latency across different batch sizes, we construct a Pareto frontier (see Figure \ref{fig:energy_latency}). Overall, the frontier exhibits a convex shape, indicating that increases in latency can lead to reductions in energy consumption, consistent with the findings of \citet{chung_mlenergy_2025}. To clarify this trade-off, we compute the percentage difference between the highest and lowest energy values observed for each LLM across batch sizes, along with the corresponding percentage difference in latency. In most cases, the percentage increase in latency exceeds the percentage decrease in energy. For example, reducing the energy consumption of Qwen 2.5-1.5B by 81\% through larger batch sizes results in a latency increase of more than 400\%. Furthermore, the magnitude of these percentage differences decreases as model size grows, suggesting that batch size variation has a smaller impact on larger models. \textbf{Therefore, careful batch size selection is essential for small-scale models, as lower energy often comes at the cost of much higher latency, a trade-off that diminishes for larger models.}

\subsection{Scaling Effects on Nano-LLMs with Additional GPUs}

\begin{figure}[ht]
  \centering
  \includegraphics[width=0.9\linewidth]{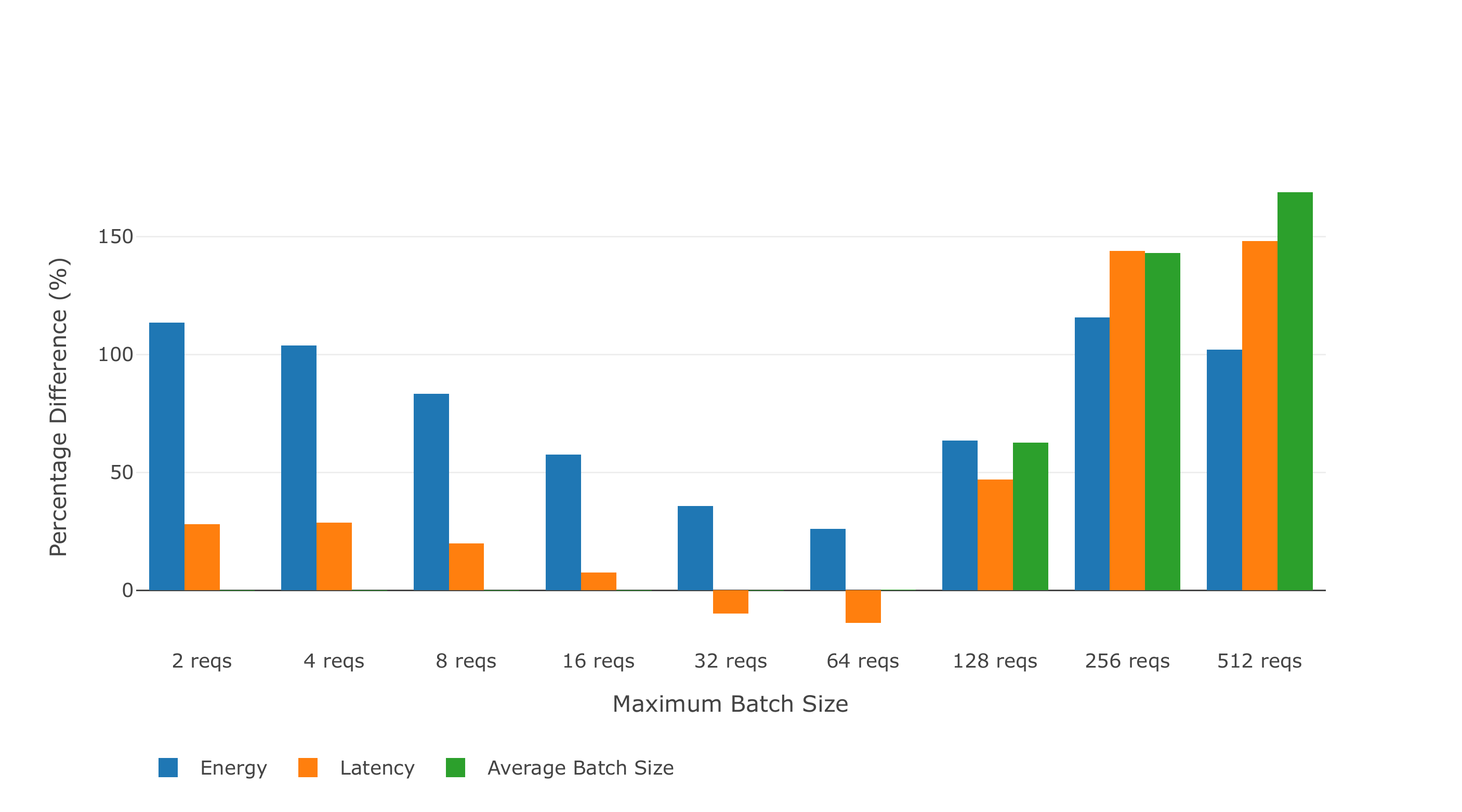}
  \caption{Percentage Difference in Energy Consumption, Decode Latency and Average Batch Size When Scaling Qwen 2.5-3B from 1 to 2 GPUs. E.g., scaling from 1 to 2 GPUs while keeping the maximum batch size fixed at 128 increases energy usage by 63\%, latency by 46\%, and average batch size by 62\%.}
%  \Description{Percentage Difference in Energy Consumption, Decode Latency and Average Batch Size When Scaling Qwen 2.5-3B from 1 to 2 GPUs. E.g., scaling from 1 to 2 GPUs while keeping the maximum batch size fixed at 128 increases energy usage by 63\%, latency by 46\%, and average batch size by 62\%.}
  \label{fig:batch_size}
\end{figure}

Nano-LLMs can typically run on a single GPU. In this experiment, we investigate their behavior when an additional GPU is introduced. Figure \ref{fig:batch_size} illustrates the percentage differences in energy consumption, decode latency, and average batch size when scaling Qwen 2.5-3B from one to two GPUs. Two key parameters are considered: maximum batch size, which represents the upper limit of requests that can fit into GPU memory (as configured in vLLM), and average batch size, which reflects the actual number of requests processed given memory constraints.

Our results (see Figure \ref{fig:batch_size}) show that increasing GPU memory does not affect the average batch size when the maximum batch size is set to 64 requests or fewer. This indicates that a single GPU is sufficient to store both model weights and the KV cache for up to 64 requests. For larger maximum batch sizes (>128 requests), the additional memory provided by the second GPU allows more requests to be processed concurrently. Below the 64-request threshold, as the maximum batch size increases, the negative impact of adding an extra GPU diminishes, implying less wasted memory and energy. Near this threshold, latency can even decrease when using two GPUs. Beyond the threshold, however, processing more requests requires additional energy and results in higher decode latency. This threshold can be estimated by calculating the memory required for the KV cache on a single GPU. \textbf{Based on these findings, we recommend tailoring GPU allocation for each model size according to request frequency and the GPU’s KV cache capacity.}

\section{Conclusion}
This study proves that smaller-scale open-weights LLMs can offer a compelling balance between sustainability and performance in agentic AI systems. By systematically evaluating a diverse set of models across energy consumption, decode latency, and output quality, we show that downsizing model size does not necessarily entail a proportional loss in performance. Our findings highlight the importance of considering deployment context, batch size optimization, and hardware configuration when selecting models for real-world applications. Furthermore, this report outlines a reproducible approach for applying these optimization techniques across different AI systems. While compression methods such as quantization and knowledge distillation can further reduce energy usage, their impact on output quality and decode latency remains variable, highlighting the need for rigorous benchmarking. However, this current work has several limitations:
\begin{itemize}
\item The benchmark centers on a specific agentic task (i.e., hallucination detection via a validation agent). While this is a representative use case, the conclusions may not fully extend to other agentic behaviors such as planning, tool use, or multi-turn reasoning. The dataset used for benchmarking is derived from a specific enterprise deployment, with long prompts and short responses. This may not reflect the diversity of real-world use cases, especially those involving different prompt lengths, languages, or domains. To address these limitations, future work should incorporate more real-world datasets and traditional benchmarks such as those used in \citet{openai_gpt-oss-120b_2025}, which cover a broader range of tasks, domains, and prompt lengths with consistent evaluation methodologies. For context-specific applications, we recommend reproducing the proposed benchmark.
\item Our comparison centers on open-weights (locally hosted) LLMs versus closed-source (API-based) models. For GPT-4o, energy consumption and latency are estimated from prior research because its architecture and serving framework are not publicly disclosed. Updating these efficiency estimates for closed-source models remains an important direction for future work. 
\item Locally hosted LLMs generally experience higher latency than API-based models, primarily because the latter have benefited from extensive optimization efforts. To close this gap, self-hosted solutions require targeted improvements. A practical starting point involves hardware upgrades, such as deploying more powerful GPUs (e.g., H100; \citet{andersch_nvidia_2022}). Beyond hardware, software-level optimizations can further reduce latency, including fine-tuning the vLLM framework, or adopting alternative LLM-serving solutions (e.g., TensorRT-LLM). For more substantial gains, advanced strategies should be considered, such as speculative decoding, disaggregated serving, and integrated software-hardware stacks, as demonstrated by \citet{elsworth_measuring_2025}.
\item The use of LLMs for response evaluation (via Ragas) can introduce potential biases and lacks the objectivity of human annotation. The absence of human evaluation may affect the reliability of quality assessments. Moreover, while our proposed metrics capture key aspects of environmental impact, user experience, and output quality, further research should explore complementary metrics (e.g., trustworthiness in output quality and peak memory usage, time between tokens, etc.) used by \citet{yang_llmcbench_2024, patel_splitwise_2024, maliakel_investigating_2025}.
\end{itemize}

Our experiments show that prompts must be fine-tuned for each LLM, as they directly affect inference energy consumption and decoding latency. The optimal prompt should be concise, reducing memory and energy usage during the pre-fill stage, while also encouraging concise output generation. Consequently, prompt optimization will be a key focus of our research direction. In addition, current experiments were conducted on AWS infrastructure and A100 GPUs. The choice of cloud service and hardware configuration (CPU, GPU) will be considered in future studies, as these factors can significantly influence energy use, $CO_2$ emissions, and water consumption.
In conclusion, this work contributes to the growing body of research identifying sustainable AI practices. It offers actionable insights for practitioners seeking to deploy scalable, efficient, and environmentally responsible AI agents without compromising user experience or system integrity.

%\printbibliography

\bibliography{bibliography}
\appendix

\section{Energy-Optimal Configurations for Experimented LLMs}

\begin{center}
	\begin{longtable}{|l||c|c|}
	\caption{Required GPUs and Energy-Optimal Maximum Batch Size for Deploying Each Large Language Model. Results correspond to the configuration with the lowest energy consumption per request across tested batch sizes (2, 4, 8, 16, 32, 64, 128, 256, 512).}
	\label{tab:result_gpu}\\
		\hline
		\textbf{LLMs} & \textbf{Number of GPUs} & \textbf{Maximum Batch Size} \\
		\hline
		\hline
		\multicolumn{3}{|l|}{\textbf{Qwen 2.5}} \\
		Qwen/Qwen2.5-0.5B-Instruct & 1 & 32 \\
		Qwen/Qwen2.5-0.5B-Instruct & 2 & 64 \\
		Qwen/Qwen2.5-1.5B-Instruct & 1 & 128 \\
		Qwen/Qwen2.5-3B-Instruct   & 1 & 128 \\
		Qwen/Qwen2.5-3B-Instruct   & 2 & 64 \\
		Qwen/Qwen2.5-7B-Instruct   & 1 & 64 \\
		Qwen/Qwen2.5-14B-Instruct  & 1 & 8 \\
		Qwen/Qwen2.5-14B-Instruct  & 2 & 256 \\
		Qwen/Qwen2.5-32B-Instruct  & 4 & 64 \\
		Qwen/Qwen2.5-72B-Instruct  & 8 & 64 \\
		\hline
		\multicolumn{3}{|l|}{\textbf{Qwen 3}} \\
		Qwen/Qwen3-4B-Instruct-2507      & 1 & 32 \\
		Qwen/Qwen3-8B                    & 1 & 32 \\
		Qwen/Qwen3-14B                   & 2 & 512 \\
		Qwen/Qwen3-30B-A3B-Instruct-2507 & 4 & 128 \\
		Qwen/Qwen3-32B                   & 4 & 512 \\
		\hline
		\multicolumn{3}{|l|}{\textbf{Distilled variants of Qwen using DeepSeek-R1}} \\
		deepseek-ai/DeepSeek-R1-Distill-Qwen-1.5B & 1 & 256 \\
		deepseek-ai/DeepSeek-R1-Distill-Qwen-7B   & 1 & 64 \\
		deepseek-ai/DeepSeek-R1-0528-Qwen3-8B     & 1 & 256 \\
		deepseek-ai/DeepSeek-R1-Distill-Qwen-14B  & 2 & 64 \\
		\hline
		\multicolumn{3}{|l|}{\textbf{Gemma 3}} \\
		google/gemma-3-4b-it   & 1 & 32 \\
		google/gemma-3-12b-it  & 2 & 16 \\
		google/gemma-3-27b-it  & 4 & 32 \\[2mm]
		\hline
		\multicolumn{3}{|l|}{\textbf{Llama 4}} \\
		meta-llama/Llama-4-Scout-17B-16E-Instruct & 8 & 128 \\
		\hline
		\multicolumn{3}{|l|}{\textbf{Phi 4}} \\
		microsoft/Phi-4-reasoning      & 2 & 256 \\
		microsoft/Phi-4-reasoning-plus & 2 & 256 \\
		\hline
		\multicolumn{3}{|l|}{\textbf{Mistral}} \\
		mistralai/Mistral-Nemo-Instruct-2407         & 2 & 32 \\
		mistralai/Mistral-Small-24B-Instruct-2501     & 4 & 256 \\
		mistralai/Mistral-Large-Instruct-2411         & 8 & 512 \\
		\hline
		\multicolumn{3}{|l|}{\textbf{Falcon 3}} \\
		tiiuae/Falcon3-1B-Instruct  & 1 & 512 \\
		tiiuae/Falcon3-3B-Instruct  & 1 & 256 \\
		tiiuae/Falcon3-7B-Instruct  & 1 & 256 \\
		tiiuae/Falcon3-10B-Instruct & 1 & 16 \\[2mm]
		\hline
		\multicolumn{3}{|l|}{\textbf{Quantized variants of Qwen 2.5 7B}} \\
		Qwen/Qwen2.5-7B-Instruct-AWQ        & 1 & 512 \\
		Qwen/Qwen2.5-7B-Instruct-GPTQ-Int4  & 1 & 256 \\
		Qwen/Qwen2.5-7B-Instruct-GPTQ-Int8  & 1 & 512 \\
		\hline
	\end{longtable}
\end{center}

\section{Output Quality versus Energy Consumption for Compressed Models}

\begin{figure}[ht]
  \centering
  \includegraphics[width=0.9\linewidth]{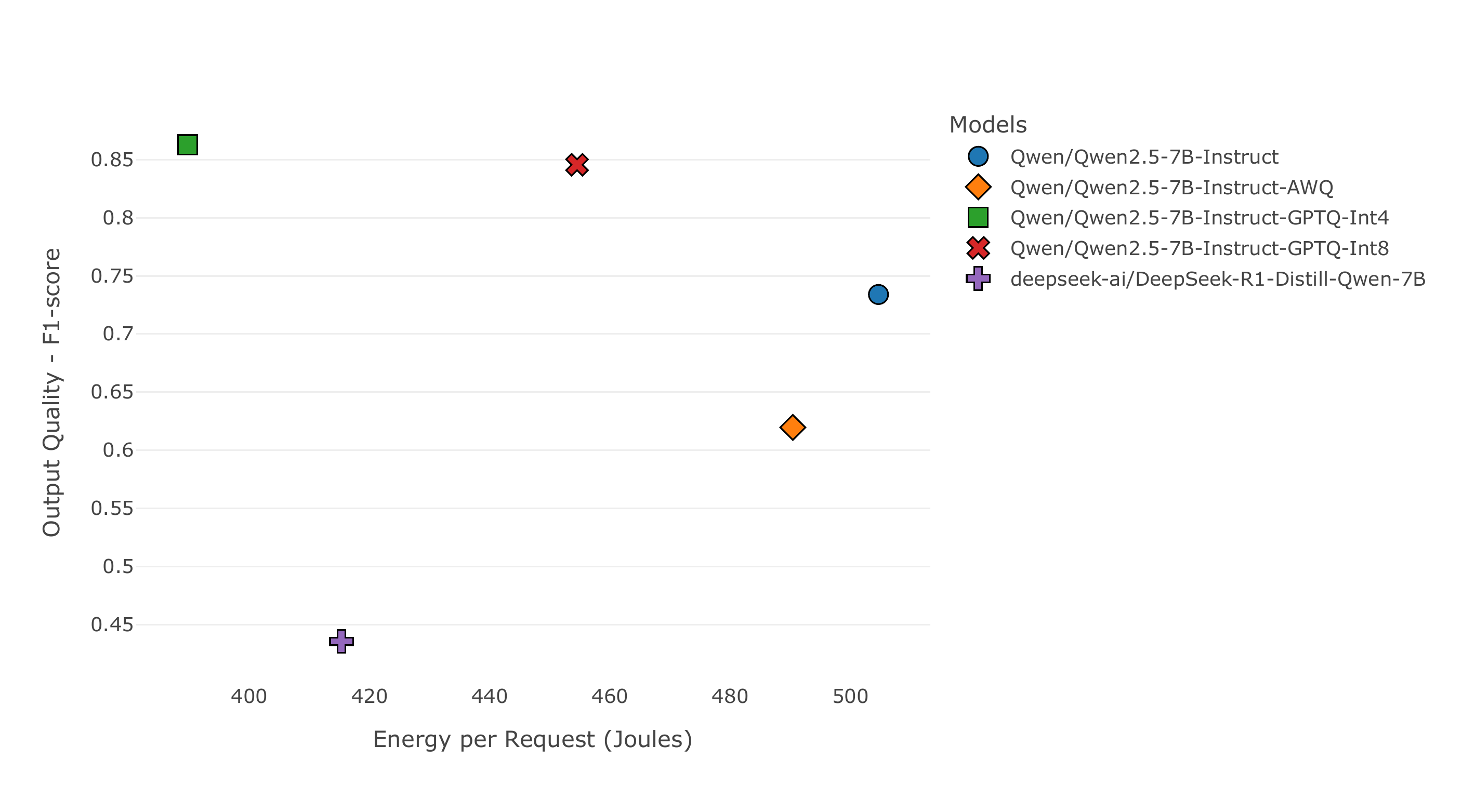}
  \caption{Output Quality (F1-score) and Energy Consumption for Compressed Qwen 2.5 7B Instruct Models. Scores correspond to the lowest-energy configuration among tested batch sizes.}
%  \Description{Output Quality (F1-score) and Energy Consumption for Compressed Qwen 2.5 7B Instruct Models. Scores correspond to the lowest-energy configuration among tested batch sizes.}
  \label{fig:compressed}
\end{figure}

\end{document}